\documentclass{article}
\usepackage{PRIMEarxiv}
\usepackage[utf8]{inputenc} 
\usepackage[T1]{fontenc}    
\usepackage{hyperref}       
\usepackage{url}            
\usepackage{booktabs}       
\usepackage{amsfonts}       
\usepackage{nicefrac}       
\usepackage{microtype}      
\usepackage{lipsum}
\usepackage{cite}
\usepackage{url}
\usepackage{fancyhdr}       
\usepackage{graphicx}       
\graphicspath{{media/}}     
\usepackage{algorithm}
\usepackage{algorithm}
\usepackage{algpseudocode}
\usepackage{amsmath}
\usepackage{tikz}
\usetikzlibrary{shapes.geometric, arrows.meta, positioning, fit}

\tikzstyle{process} = [rectangle, rounded corners, minimum width=2.8cm, minimum height=1cm,text centered, draw=black, fill=purple!20]
\tikzstyle{data} = [cylinder, cylinder uses custom fill, cylinder body fill=blue!20, cylinder end fill=blue!10, shape border rotate=90, aspect=0.25, draw, minimum height=1cm, minimum width=1cm]
\tikzstyle{arrow} = [thick,->,>=stealth]
\tikzstyle{smallbox} = [rectangle, draw=black, fill=orange!20, text width=3cm, minimum height=1cm, text centered]

\title{Small Sample-Based Adaptive Text Classification through Iterative and Contrastive Description Refinement}

\author{
  Amrit Rajeev, Udayaadithya Avadhanam, Harshula Tulapurkar, Sai Barath Sundar   \\
  Mphasis Limited\\
  \texttt{\{amrit.r1, udayaadithya.a, harshula.tulapurkar, sai.sundar\}@mphasis.com} \\
}

\begin{document}
\maketitle

\begin{abstract}
Zero-shot text classification remains a difficult task in domains with evolving knowledge and ambiguous category boundaries, such as ticketing systems. Large language models (LLMs) often struggle to generalize in these scenarios due to limited topic separability, while few-shot methods are constrained by insufficient data diversity. We propose a classification framework that combines iterative topic refinement, contrastive prompting, and active learning. Starting with a small set of labeled samples, the model generates initial topic labels. Misclassified or ambiguous samples are then used in an iterative contrastive prompting process to refine category distinctions by explicitly teaching the model to differentiate between closely related classes. The framework features a human-in-the-loop component, allowing users to introduce or revise category definitions in natural language. This enables seamless integration of new, unseen categories without retraining, making the system well-suited for real-world, dynamic environments. The evaluations on AGNews and DBpedia demonstrate strong performance: 91\% accuracy on AGNews (3 seen, 1 unseen class) and 84\% on DBpedia (8 seen, 1 unseen), with minimal accuracy shift after introducing unseen classes (82\% and 87\%, respectively). The results highlight the effectiveness of prompt-based semantic reasoning for fine-grained classification with limited supervision.
\end{abstract}

\keywords{Zero-shot learning, text classification, contrastive prompting, active learning, iterative refinement, large language models, natural language processing }

\section{Introduction}
Zero-shot text classification remains a complex challenge, particularly in domains with evolving categories and ambiguous boundaries, such as ticketing systems, knowledge base tagging, and customer support. In these environments, LLMs face difficulties in generalizing due to limited topic separability and the fluid nature of category definitions. While supervised learning models like .\cite{meng2020text} which associates semantically related words with the label names to self-train the model performs well in controlled settings, they rely on the assumption that label names are semantically rich and representative of their respective categories, which are not always feasible for real-world applications.

Moreover, training based methods struggle to adapt to new categories without retraining the entire model. Recent research like PESCO .\cite{wang2023pesco}, prompt-learning for short text classification .\cite{zhu2023prompt} and few-shot learning \cite{meng2020text} with LLMs have shown promising results for text classification However, these approaches often fail to capture the full complexity of real-world classification scenarios, particularly when dealing with overlapping categories or achieving reasonable performance in cost effective way.  For instance, while prompt-learning techniques have shown efficacy in short text classification, they may incur omissions and biases due to manual label word expansion or limited consideration of class names. Similarly, few-shot learning approaches, though effective in low-resource settings, can face challenges in scalability and performance degradation as the number of novel classes increases.  

Prompt engineering and instruction tuning methods have shown promise in leveraging the contextual abilities of LLMs for zero- and few-shot learning. However, these techniques also struggle with classes that are semantically similar or overlapping. For example, Generation-Driven Contrastive Self-Training (GenCo) \cite{zhang2023generation} uses an instruction-following GPT model to enhance a sentence encoder through self-training. While it introduces useful mechanisms like input and conditional augmentation, its computational cost remains high, as multiple GPT queries are required per iteration, making it impractical for real-time applications. 

Several methodologies have been proposed to also address the challenge of data scarcity in text classification tasks. Pattern-Exploiting Training (PET) \cite{schick2020exploiting} represents a notable approach that significantly enhances the capabilities of pre-trained language models in low-resource scenarios. Here, instead of directly fine-tuning models on labeled examples, PET reformulates classification tasks into close-style (fill-in-the-blank) sentences using manually crafted patterns. These transformed inputs are used to fine-tune language models on a small dataset, and their predictions are then used to label a much larger unlabeled dataset with soft labels. A final classifier is trained on this pseudo-labeled data. However, it is highly dependent on the quality and manual design of the patterns and verbalizers, making it less scalable across tasks or domains; it also requires training multiple models per pattern, increasing computational cost.

Despite the potential of these techniques, many of them treat the model as static, requiring retraining to incorporate new classes. To address this gap, we propose a novel classification framework that combines iterative topic refinement, contrastive prompting, and active learning. Instead of treating classification as a task over labeled instances, we frame it as a semantic reasoning process over natural language descriptions. The model begins with a small set of labeled samples (n=20) to generate initial topic labels, which are then iteratively refined through contrastive prompting, teaching the model to distinguish between closely related categories. Importantly, our framework adapts seamlessly to new, unseen categories, enabling real-time integration without the need for retraining. This makes it especially suitable for dynamic environments where categories are continuously evolving.

\section{Related work}
\label{sec:headings}

\subsection{Generalized zero-shot text classification}

Generalized zero-shot text classification represents a complex task that involves the categorization of textual content across two distinct domains: previously encountered classes and newly emerging, unseen categories. Current methodologies exhibit significant limitations in their generalization capabilities, primarily due to their parameter optimization being exclusively tailored to known classes rather than accommodating both seen and unseen categories. Furthermore, these approaches suffer from rigidity in their predictive mechanisms, as their parameters remain static during the classification process. This inherent inflexibility undermines their ability to adapt to novel categories, resulting in suboptimal performance when confronted with previously unseen classes. Even methods like Learn to Adapt for Generalized Zero-Shot Text Classification \cite{zhang2022learn} involve  significant training of the model. 

\subsection{BYOC: Personalized few-shot classification with co-authored class descriptions}

BYOC (Bring Your Own Categories) \cite{bohra2023byoc} introduces a human-in-the-loop framework in which users label a small set of examples and answer LLM-generated questions to incrementally refine category descriptions. Although successful in low-data settings, BYOC focuses on single-turn refinement and lacks mechanisms for distinguishing between semantically similar classes. Our approach extends this idea by integrating both iterative refinement and contrastive prompting in a unified pipeline, enabling more precise category boundary formation over multiple learning cycles.

\subsection{Active learning and human feedback}

Several recent studies have explored active learning with LLMs to reduce annotation cost. PATRON \cite{yu2022cold} and Active Few-Shot Learning \cite{ahmadnia2025active} focus on selecting the most informative samples for annotation based on model uncertainty and diversity. However, they do not incorporate user input for semantic refinement. Other frameworks, such as Enhancing Text Classification through LLM-Driven Active Learning and Human Annotation \cite{rouzegar2024enhancing}, involve humans in labeling uncertain examples but treat label definitions as fixed. Our method differs by allowing users to dynamically modify and create new class descriptions throughout the process, merging active selection with semantic evolution.

\section{Problem formulation}

Given a set of documents D = $\{x_1, x_2, ..., x_n\}$ and a predefined set of categories C = $\{c_1, c_2, ..., c_z\}$, our objective is to develop a robust zero-shot classification framework that can effectively categorize documents with minimal labeled data while maintaining high accuracy. The problem can be formally decomposed into several interconnected components: 

\begin{enumerate}

    \item \textbf{Category description generation} Let $S$ be a function that generates semantic descriptions for each category. For each category $c_i \in C$: $S(c_i) = d_i$, where $d_i$ represents the natural language description of category $c_i$. This description should maximize the discriminative power between categories while maintaining semantic coherence.

    \item \textbf{Contrastive learning objective} To enhance the model's ability to distinguish between similar categories.

    \item \textbf{Iterative refinement process} The description refinement process is formulated as an iterative optimization problem. See Algorithm \ref{alg:refine}

    \begin{algorithm}[ht]
    \caption{Algorithm for refinement step for a category $c_0$}
    \label{alg:refine}
    \begin{algorithmic}[1]
    \Require Initial description $d_{00}$ for category $c_0$, input document sampler $D(t)$, maximum iterations $T$
    \Ensure Refined description $d_{0r}$ for category $c_0$
    \Statex
    \For{$t \gets 0$ to $T - 1$}
        \State $d_{0t+1} \gets$ \Call{Refine}{$d_{0t}$, $D(t)$} \Comment{Apply refinement function with current description and document sample}
    \EndFor
    \State $d_{0r} \gets d_{0T}$
    \State \Return $d_{0r}$
    \end{algorithmic}
    \end{algorithm}

    \item \textbf{Classification function} The classification task is formulated as: 

    \begin{algorithm}[ht]
    \caption{Algorithm for classification}
    \label{alg:classify}
    \begin{algorithmic}[1]
    \Require Input document $x$, categories $C = \{c_1, c_2, ...., c_n\}$, refined descriptions for all categories $D = \{d_{0r}, d_{1r}, ..., d_{nr}\}$, Prompt template for classification task $I$, decoder-based LLM that outputs probabilities over tokens $f_{LLM}$
    \Ensure Predicted category $c_{pred}$
    \Statex
    \State $\hat{c} \gets \arg\max_{c_i \in \mathcal{C}} P_{\text{LLM}}(c_i \mid I(x, D))$ \Comment{The LLM computes the likelihood of each label $c_i$
  as the decoder output probability of generating the label $c_i$ immediately after the prompt.}
    \Statex
    \State $c_{pred} \gets C[\hat{c}]$
    \State \Return $c_{pred}$
    \end{algorithmic}
    \end{algorithm}

\end{enumerate}

\section{Methodology}

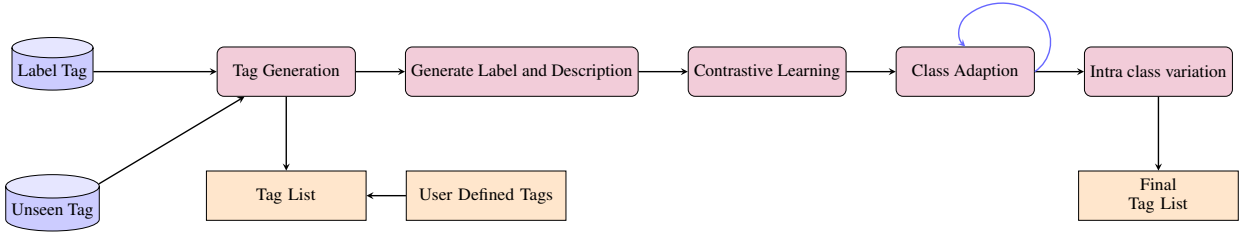
\begin{figure}
    \resizebox{\textwidth}{!}{
    \begin{tikzpicture}[node distance=1.7cm and 1cm]
    
    \node (labelTag) [data] {Label Tag};
    \node (unseenTag) [data, below=of labelTag] {Unseen Tag};
    
    \node (tagGen) [process, right=2.5cm of labelTag] {Tag Generation};
    \node (genLabel) [process, right=of tagGen] {Generate Label and Description};
    \node (contrast) [process, right=of genLabel] {Contrastive Learning};
    \node (classAdopt) [process, right=of contrast] {Class Adaption};
    \node (intraVar) [process, right=of classAdopt] {Intra class variation};
    
    \node (tagList) [smallbox, below=1.5cm of tagGen] {Tag List};
    \node (userTags) [smallbox, right=0.8cm of tagList] {User Defined Tags};
    
    \node (finalTag) [smallbox, below=1.5cm of intraVar] {Final\\ Tag List};
    
    \draw [arrow] (labelTag) -- (tagGen);
    \draw [arrow] (unseenTag) -- (tagGen);
    
    \draw [arrow] (tagGen) -- (genLabel);
    \draw [arrow] (genLabel) -- (contrast);
    \draw [arrow] (contrast) -- (classAdopt);
    \draw [arrow] (classAdopt) -- (intraVar);
    
    \draw [arrow] (tagGen) -- (tagList);
    \draw [arrow] (userTags) -- (tagList);
    
    \draw [arrow] (intraVar) -- (finalTag);
    
    \draw [arrow, bend right=45, blue!60, thick] (classAdopt.east) to ([xshift=0.2cm, yshift=1.0cm]classAdopt.east) to ([yshift=0.6cm]classAdopt.north) to (classAdopt.north);
    
    \end{tikzpicture}
    }
    \caption{Flow diagram}
    \label{fig:enter-label}
\end{figure}

Our framework is a self-optimizing and evolving pipeline that leverages zero-shot capabilities of LLMs as represented in Fig.~\ref{fig:enter-label}.

This framework involves five steps:

\subsection{Generate pseudo labels and description}

In our approach, we deliberately move away from using original labels due to the inherent subjectivity in human labeling processes which can lead to inconsistent labels across similar data points.   Consider the following example from a customer support ticketing system:

Original Ticket: "Unable to login to the application after the recent update. Keep getting 'Invalid Credentials' error even though I'm sure my password is correct."

This ticket could be labeled differently by different annotators:
•	Annotator 1: "Login Issues"

•	Annotator 2: "Authentication Error"

•	Annotator 3: "Post-Update Problems"

•	Annotator 4: "Password Problems"

While all these labels are technically correct, they capture different aspects of the same issue, leading to inconsistency in the training data.  Instead, our approach takes N similar samples from a category which can be as low as 20. The samples you pick can be random or the first 20 samples. Next, we generate a comprehensive semantic description using LLM that captures the underlying theme of the topic. For the login-related issues above, our system might generate is given in A1 of appendix.

This generated description provides a more holistic and consistent representation of the category, encompassing various manifestations of authentication-related issues. By using this approach, we:

•	Eliminate individual annotator biases

•	Create more comprehensive category definitions

•	Capture the fundamental essence of the category rather than surface-level labels

•	Provide clear boundaries for classification while maintaining flexibility for similar cases

The system generates these topic-description pairs by analyzing patterns and common themes across the 20 sample instances, ensuring that the resulting description is both specific enough to be meaningful and broad enough to capture various intricacies of the same underlying  issue.

\subsection{Contrastive learning}

Contrastive learning was introduced in \cite{chopra2005learning} where the idea is to place similar data points together. Inspired by PESCO’s method of maximizing the similarity between input and its label \cite{wang2023pesco} we employ contrastive learning mechanisms, which enhances the model's ability to discriminate between semantically similar categories. This will inversely help the model learn the similarity between our input description and its correct category.While the initial topic descriptions provide a foundation for classification, real-world categories often have subtle overlaps that can lead to misclassifications. Our contrastive learning step addresses this by explicitly teaching the model to understand the nuanced differences between related categories. 

 For instance, if two categories like "Software\_Authen- tication\_Access\_Issues" and "Hardware\_Authenticat- ion\_Access\_Issues" show high similarity, the system generates descriptions that emphasize their unique characteristics, an example for this given in appendix A2.

 This generated description provides a more holistic and consistent representation of the category.  By incorporating category descriptions into the classification process, models are guided to focus on the most salient features pertinent to each category, enhancing their ability to distinguish between nuanced classes.

\subsection{Fine-grained discriminative refinement}

The description refinement phase represents a crucial iterative process in our methodology that continuously enhances category descriptions based on real-world classification performance. After establishing initial descriptions from the first N samples, we proceed with a validation phase using an additional M data points per category. The selection of parameter M can be calibrated according to the available data volume. However, the central thesis of this research is to demonstrate efficacy in low-data scenarios. Consequently, we deliberately constrained our experimental design to 40-50 samples, which represents a minimalist yet sufficient dataset to validate our approach.

 This step serves as both a performance check and a refinement mechanism. When the model's classification accuracy falls below say an 80\% threshold on the M samples (i.e., more than 20\% misclassifications) for any category, we trigger an automated refinement process. An example for this is given in appendix A 3.1.

 Upon analyzing misclassified cases, let’s say the system might identify additional patterns or scenarios it failed to capture, such as VPN-related issues or proxy server problems. The description is then automatically augmented to incorporate these missing concepts. An example for this is given in appendix A 3.2

 This refinement process is particularly powerful as it learns from actual classification errors, ensuring that the descriptions evolve to cover edge cases and previously unconsidered scenarios. The iterative optimization process terminates upon either achieving a classification accuracy threshold exceeding 80\% on the designated refinement dataset or upon completion of four (though this parameter remains adjustable, our empirical observations indicate performance stabilization typically occurs after four iterations, with subsequent modifications yielding negligible improvement in classification metrics) optimization epochs, after which the system transitions to the subsequent phase outlined in Section D. The iterative nature of this approach allows the system to continuously improve its understanding of each category, leading to more robust and comprehensive classification capabilities. Importantly, this refinement occurs without requiring complete retraining or extensive manual intervention, making it both efficient and practical for real-world applications.

\subsection{Class adaption}

Despite the initial topic refinement process described in section 4.3, challenges can persist in distinguishing between semantically similar classes. A notable example in this can be the frequent misclassification between "Software\_Authent- ication\_Access\_Issues" and "Hardware\_Authenticati- on\_Access\_Issues". This confusion arises even when the classes have distinct descriptors, as illustrated in the description after section 4.3 which is as shown in A4.1 appendix 

To address this limitation, we introduce an intra-class adaptation mechanism that specifically targets the fine-grained distinctions between closely related classes. This approach involves analyzing the misclassified instances to identify the subtle but critical differences between similar categories. Within this phase, we inference the model and elicit the misclassified categorical pairs which are systematically introduced into the refinement pipeline, wherein the Large Language Model (LLM) engages in targeted description enhancement of the misclassified categorical pairs. This process specifically addresses the fine-grained distinguishing characteristics that differentiate between commonly conflated category pairs. Through this discriminative refinement approach, the model develops increased sensitivity to subtle taxonomic boundaries, thereby improving classification precision in areas where categorical overlap previously led to classification errors.

 The updated description after the refinement can look like A4.2

\subsection{New topic/Custom topic adaption}

Recent approaches in zero-shot classification have attempted to address inter-class relationships through sophisticated prototype construction for unseen classes \cite{liu2019reconstructing}\cite{schick2020exploiting}. While these methods show promise, they typically require extensive embedding training and substantial computational resources. Our work diverges from these resource-intensive approaches, instead of proposing a more efficient and cost-effective method for learning class contrasts through natural language understanding. We introduce a dynamic adaptation mechanism that effectively handles evolving classification requirements without the need for embedding training or model retraining. This is particularly crucial in real-world applications where new categories frequently emerge and need to be seamlessly integrated into the existing classification system. Our framework implements this through a structured adaptation process that leverages the inherent language understanding capabilities of LLMs.

\section{Experiments and results}

To rigorously assess the efficacy of our proposed approach, we conducted a comprehensive comparative analysis against several state-of-the-art models and methodologies. Specifically, our evaluation compared our framework against Llama 3.1 8B \cite{grattafiori2024llama}, IReRa \cite{d2024context} and Pesco \cite{wang2023pesco}. The comparative performance metrics are presented in Table 1, providing a quantitative basis for evaluating our method's relative advantages. To evaluate the effectiveness of our proposed classification framework, we primarily utilize two widely adopted datasets: AG News \cite{d2024context} and DBPedia. The AG News dataset comprises news articles categorized into four distinct classes—World, Sports, Business, and Science/Technology—providing a balanced benchmark for evaluating topic-level text classification. In contrast, DBPedia consists of ontology-based Wikipedia entries labeled across 14 fine-grained categories.

\begin{table}[h]
\centering
\caption{Results Comparison}
\label{tab:results}
\begin{tabular}{|l|c|c|c|c|}
\hline
\textbf{Method} & \textbf{AGNews Seen} & \textbf{AGNews Unseen} & \textbf{DBPedia Seen} & \textbf{DBPedia Unseen} \\
\hline
Llama 3.1 8B Zero Shot & 83.6\% & 81\% & 76\% & 80\% \\
\hline
IReRa & 82.5\% & 78\% & 57\% & 63\% \\
\hline
Pesco & 90.67\% & 79\%* & 86\% & 72\%* \\
\hline
\textbf{Ours} & \textbf{91.1\%} & \textbf{83\%} & \textbf{84\%} & \textbf{87\%} \\
\hline
\end{tabular}
\vspace{1ex}

\textsuperscript{*}Note: For Pesco, which is a training algorithm, we increased the sample size of the unseen class set to 1000 instances to obtain comparable results. All other models were evaluated using identical experimental conditions as described in our methodology.
\end{table}

We conducted extensive experimentation by systematically varying key sampling parameters across multiple configurations. We implemented a controlled study wherein the value of N (the number of initial samples per class) was varied across a meaningful range. For each N value, we generated multiple random subsets from the available training data to ensure our findings were not artifacts of specific sample selections. This approach allowed us to observe the pipeline's stability across different initialization conditions.

 Similarly, we examined the impact of varying M (the number of samples used during the refinement phase). This investigation helped determine the optimal amount of data required for effective refinement without introducing unnecessary computational overhead.

 Our experimentation revealed that the pipeline exhibits remarkable stability across different sampling configurations. Performance metrics remained consistent within a narrow margin across various N and M values, suggesting that our approach is not overly sensitive to the precise number of examples used during initialization and refinement.

 Despite this general stability, we identified one significant limitation: when the number of initial samples becomes excessively large, the generated class descriptions tend to become overly broad. This broadening effect occurs because the system attempts to accommodate the increased variability present in larger sample sets.

The overly generalized descriptions can subsequently impair the precision of the tag assignment process, as the distinctive characteristics that effectively differentiate between classes become diluted. This finding suggests an important design consideration: while additional examples generally improve machine learning systems, our pipeline benefits from focused, representative samples rather than exhaustive enumeration during the description generation phase.

This trade-off between sample size and description specificity represents an important calibration point when deploying our system in practical applications.

\section{Limitations}

Although our approach significantly reduces the amount of labeled data required compared to traditional methods, the quality of the initial sample remains critical. The system's performance is sensitive to the representativeness of the small subset (n=20) used for generating initial tags and descriptions. If these samples fail to capture key class characteristics, subsequent refinement may build upon an inadequate foundation.

 As the number of categories increases, the contrastive prompting mechanism faces increased complexity. The system must maintain distinctions between an expanding set of potentially similar categories, which may exceed the effective context window of the underlying LLM. This limitation becomes particularly apparent in taxonomies with hundreds of fine-grained categories.

 The iterative refinement process, while effective, requires multiple LLM calls per refinement cycle. This introduces computational overhead that may be prohibitive in resource-constrained environments or when dealing with very large datasets that require real-time classification.

\bibliographystyle{unsrt}  
\bibliography{references}

\appendix
\section*{Appendix A: Examples}

A1 Example for generated sudo labels  

\{ "topic\_name": "Authentication\_Access\_Issues", 
    "topic\_description": "Issues related to system access, including login failures, authentication errors, and credential validation problems. This encompasses situations where users cannot access their accounts due to authentication mechanisms, regardless of whether the issue stems from password mismatches, system updates, or session management problems." 
\}. 

A2 Example output for contrast learning  

\{ 

 "topic\_name": " Software\_Authentication\_Access\_Issues ", 

 "topic\_description": "Issues related to A B C and D. This is different from other classes because it doesn’t contain issues pertaining to X Y Z or Oracle login issues. " 

\} 

\{ 

"topic\_name": " Hardware\_Authentication\_Access\_Issues", 

"topic\_description": " Issues related to X Y and Z. This is different from other classes because it doesn’t contain issues pertaining to A, B, C and D or Oracle login issues. 

\} 

\{ 

 "topic\_name": " Oracle\_Authentication\_Access\_Issues ", 

 "topic\_description": "Issues related specifically to oracle login and does not involve issues related to A B C D X Y and Z " 

\}.

A3 

A3.1  

Examples for class adaptation.

Initial Description: 

\{ 

"topic\_name": "Network\_Connectivity", 

 "topic\_description": "Issues related to network connection failures, including internet outages and connection drops."

\}

A3.2   

Refined Description: 

\{ "topic\_name": "Network\_Connectivity", 

"topic\_description": "Issues related to network connection failures, including internet outages, connection drops, VPN access problems, proxy server connectivity, and network authentication failures. This encompasses both complete connection loss and intermittent connectivity issues across various network configurations." 

\}

A4 

A4.1 The description after section 2.3

\{  

"topic\_name": "Software\_Authentication\_Access\_Issues", 

"topic\_description": "Issues related to A B C and D. This is different from other classes because it doesn't contain issues pertaining to X Y Z or Oracle login issues." 

\}

\{ 

"topic\_name": "Hardware\_Authentication\_Access\_Issues", 

"topic\_description": "Issues related to X Y and Z. This is different from other classes because it doesn't contain issues pertaining to A, B, C and D or Oracle login issues." 

\}

A4.2 The description after refinement. 

\{ 

"topic\_name": "Software\_Authentication\_Access\_Issues", 

"topic\_description": "Authentication issues specifically related to A, B, and C. This includes problems with password resets, digital certificate validations, software token authentication, and user account lockouts in applications. Unlike hardware authentication issues, these problems don't involve physical access devices or biometric hardware malfunctions. Key identifiers include error messages related to invalid credentials, permission denied notifications, and software-side session timeouts." 

\} 

\{ 

"topic\_name": "Hardware\_Authentication\_Access\_Issues", 

"topic\_description": "Authentication issues specifically related to X, Y and Z. This encompasses problems with biometric scanners, security card readers, hardware security tokens, and physical access control units. Unlike software authentication issues, these problems involve tangible device malfunctions rather than digital credential issues. Key identifiers include device connectivity errors, biometric reader failures, and physical token recognition problems." 

\}

A5  

// Initial User Input 

\{ 

"topic\_name": "Cloud\_Infrastructure\_Issues", 

" topic\_description ": "This about talks about EC2 instance not launching and Azure VM connectivity failure" 

\} 

// Refined tag after section 2.4 can be 

\{ 

"topic\_name": "Cloud\_Infrastructure\_Issues", 

"topic\_description": "Problems specifically related to cloud infrastructure services and platforms. This encompasses issues with cloud resource provisioning (EC2, Azure VM), cloud networking (VPC, subnets), and service availability. Distinguished from general network issues by its focus on cloud-specific components and services. Key identifiers include cloud platform-specific error codes, instance state problems, and cloud resource allocation failures. Unlike software authentication issues, these focus on infrastructure-level problems rather than application-level access." 

\} 

\section*{Appendix B Prompts}

B1 Sample prompt for generating Tag and Description

""" You are an AI assistant specialized in analyzing data and generating concise, relevant tags. You've been given a subset of a dataframe containing statements. Your task is to:

Carefully analyze the ticket Description provided by the user.

Based on the information given:

Identify the most prominent theme or topic from the Description 

Generate a single, highly relevant tag that best summarizes the main Category or subject 

Provide a brief (2-3 sentence) Description of what this tag represents, covering the most common aspects from the Description. 

Present your analysis in a structured JSON format as follows:  
\begin{quote}
\texttt{\{\{} \\
\texttt{"topic\_name": "<topicName>",} \\
\texttt{"topic\_Description": "<topicDescription>"} \\
\texttt{\}\}}
\end{quote}

Ensure your tag is concise, relevant, and reflective of the most common Category in the Description. Your Description should explain what this tag represents and how it relates to the overall context.

Respond only with the JSON output for the given Description. Do not include any additional text or explanations outside the JSON structure."""

B2 Prompt for contrasting

"""You are an AI assistant trained to analyze data and generate concise, relevant tags. You've been given multiple subsets of wikipedia articles. Your task is to:

\begin{enumerate}
    \item Consider these tags and their initial Descriptions:

    \begin{quote}
    \texttt{\{category\}}
    \end{quote}

    \item Based on the tag Descriptions, refine all tags' Descriptions such that they emphasize the contrast between all categories to make them easily distinguishable. Your updated Descriptions should:
    
    \begin{itemize}
        \item Be 3--4 sentences long each.
        \item Capture the most common themes from the respective tag Descriptions.
        \item Accurately represent what each tag encompasses.
    \end{itemize}
\end{enumerate}

- Highlight the key differences between all categories 

    3. Present your analysis in this JSON format similar to the input.

Remember: 

        - Focus on the most prominent Categorys or topics in each subset of Descriptions 

    - Ensure your tag Descriptions are clear, concise, and relevant 

    - Aim for Descriptions that would be helpful for customer service agents categorizing future tickets 

    - Emphasize the contrast between all categories to make them easily distinguishable 

    - Use comparative language to highlight unique aspects of each category 

    - Identify any overlapping themes and explain how they differ in each category

Your response should only include the requested JSON output, without any additional explanations or comments. Ensure that the Descriptions for each category clearly differentiate it from all others, creating a comprehensive and contrasting set of tag definitions."""

B3 Inter Class Adaption

 """You are an expert system specializing in content classification and taxonomy refinement. Your task is to analyze content categorization patterns and enhance category definitions. You will be provided with: 

\begin{enumerate}
    \item \textbf{Successfully Categorized Content} (\texttt{df\_subset\_right}):
    
    \begin{quote}
    \texttt{\{df\_subset\_right\}}
    \end{quote}
    
    These items were correctly identified as belonging to \texttt{\{correct\_category\}}.
    
    \item \textbf{Miscategorized Content} (\texttt{df\_subset\_wrong}):
    
    \begin{quote}
    \texttt{\{df\_subset\_wrong\}}
    \end{quote}
    
    These items should be in \texttt{\{correct\_category\}} but were mistakenly classified as \texttt{\{wrong\_category\}}.
\end{enumerate}

Your objective is to modify the category definitions to:

1. Strengthen the distinction between \texttt{\{correct\_category\}} and \texttt{\{wrong\_category\}}. 

2. Identify unique markers and patterns in correctly classified content.

3. Analyze the characteristics that led to misclassification 

4. Establish clearer boundaries between overlapping categories 

5. Incorporate discriminative features from the correctly classified examples 

\textbf{Refinement Guidelines:}

\begin{itemize}
    \item Maintain the core meaning of each category.
    \item Add specific distinguishing characteristics.
    \item Include explicit exclusion criteria where necessary.
    \item Use precise, unambiguous language.
    \item Incorporate key differentiating keywords.
    \item Address common points of confusion.
\end{itemize}

Present your refined category in this JSON format:

\begin{quote}
\texttt{\{\{} \\
\texttt{"topic\_name": "<topicName>",} \\
\texttt{"topic\_Description": "<modifiedTopicDescription>"} \\
\texttt{\}\}}
\end{quote}

Ensure your response is well-structured, adheres to proper JSON syntax, and includes only the JSON output for the single refined category tag.

Return only the JSON object, nothing else"""

B4 Prompt used to refine description based on misclassified outputs 

"""You are an AI assistant specializing in data analysis. Your task is to refine a category tag based on correctly classified and misclassified Descriptions. Analyze the following data:

1. Correctly classified data: 

 \{df\_subset\_right\} 

 These data points belong to the \{correct\_category\} category and were correctly labeled. 

2. Misclassified data: 

 \{df\_subset\_wrong\} 

These data points belong to the \{correct\_category\} category but were incorrectly labeled. 
Based on this data, refine the \{category\_list\} by:

1. Identifying the primary themes from misclassified data. 

2. Understanding why some data points were misclassified. 

 3. Defining the category's scope and boundaries accurately. 

4. Highlighting key commonalities across all data points. 

5. Using clear, specific language to minimize future misclassifications.

Your refined category Description should: 

\begin{itemize}
    \item Do not change the original topic name.
    \item Consist of 2--3 concise, informative sentences.
    \item Capture the essence of the most frequent categories in the correct category.
    \item Clearly explain what the tag represents and its relationship to the category.
    \item Include relevant keywords to improve future classification accuracy.
    \item Address potential areas of confusion that led to the initial misclassification.
\end{itemize}

Present your refined category in this JSON format:

\begin{quote}
\texttt{\{\{} \\
\texttt{"topic\_name": "<topicName>",} \\
\texttt{"topic\_Description": "<modifiedTopicDescription>"} \\
\texttt{\}\}}
\end{quote}

Ensure your response is well-structured, adheres to proper JSON syntax, and includes only the JSON output for the single refined category tag.  

Return only the JSON object, nothing else"""

B5 Intra Class differentiation

"""You are an expert system specializing in content classification and taxonomy refinement. Your task is to analyze content categorization patterns and enhance category definitions. You will be provided with:

\textbf{Successfully Categorized Content} (\texttt{df\_subset\_right}): 

\begin{quote}
\texttt{\{df\_subset\_right\}}
\end{quote}

These items were correctly identified as belonging to \texttt{\{correct\_category\}}.

\vspace{1em}

\textbf{Miscategorized Content} (\texttt{df\_subset\_wrong}): 

\begin{quote}
\texttt{\{df\_subset\_wrong\}}
\end{quote}

These items should be in \texttt{\{correct\_category\}} but were mistakenly classified as \texttt{\{wrong\_category\}}.

Your objective is to modify the category definitions to: 
        1. Strengthen the distinction between \texttt{\{correct\_category\}} and \texttt{\{wrong\_category\}}.

        2. Identify unique markers and patterns in correctly classified content 

        3. Analyze the characteristics that led to misclassification 

        4. Establish clearer boundaries between overlapping categories 

        5. Incorporate discriminative features from the correctly classified examples

        Refinement Guidelines: 

        - Maintain the core meaning of each category 

        - Add specific distinguishing characteristics 

        - Include explicit exclusion criteria where necessary 

        - Use precise, unambiguous language 

        - Incorporate key differentiating keywords 

        - Address common points of confusion

Present your refined category in this JSON format: 

        \{\{ 
        "topic\_name": "<topicName>", 
        "topic\_Description": "<modifiedTopicDescription>" 
        /\}\} 

Ensure your response is well-structured, adheres to proper JSON syntax, and includes only the JSON output for the single refined category tag.

Return only the JSON object, nothing else""" 

\end{document}